\documentclass{article}
\usepackage{spconf,amsmath,graphicx}

\usepackage{enumitem}
\setlist{nosep, leftmargin=14pt}

\usepackage{graphicx}
\usepackage{subfigure}
\usepackage{mwe} 
\usepackage{amsfonts}
\usepackage[figuresright]{rotating}
\usepackage{longtable}
\usepackage{graphics}
\usepackage{threeparttable}
\usepackage{booktabs}
\usepackage{multirow}
\usepackage{multicol}
\usepackage{makecell}
\usepackage{adjustbox}
\usepackage[dvipsnames]{xcolor}
\usepackage{bbding}
\usepackage{diagbox}
\usepackage{pifont}
\usepackage{pdflscape}
\newcommand{\E}{\mathcal{E}}
\newcommand{\G}{\mathcal{G}}
\newcommand{\D}{\mathcal{D}}

\newcommand{\Loss}{\mathcal{L}}

\newcommand\blfootnote[1]{
    \begingroup
    \renewcommand\thefootnote{}\footnote{#1}
    \addtocounter{footnote}{-1}
    \endgroup
}

\title{Generative Medical Image Anonymization Based on Latent Code Projection and Optimization}
\name{Huiyu Li, Nicholas Ayache, Hervé Delingette}
\address{Inria, Epione Team, Université Côte d’Azur, Sophia Antipolis, France}
\begin{document}
%
\maketitle
\begin{abstract}
Medical image anonymization aims to protect patient privacy by removing identifying information, while preserving the data utility to solve downstream tasks. 
In this paper, we address the medical image anonymization problem with a two-stage solution: latent code projection and optimization. In the projection stage, we design a streamlined encoder to project input images into a latent space and propose a co-training scheme to enhance the projection process. In the optimization stage, we refine the latent code using two deep loss functions designed to address the trade-off between identity protection and data utility dedicated to medical images. 
Through a comprehensive set of qualitative and quantitative experiments, we showcase the effectiveness of our approach on the MIMIC-CXR chest X-ray dataset by generating anonymized synthetic images that can serve as training set for detecting lung pathologies. Source codes are available at https://github.com/Huiyu-Li/GMIA.

\end{abstract}
\section{Introduction}
\label{sec:intro}
Medical images often contain sensitive information that can be linked to individual patients \cite{packhauser2022deep}, posing risks of privacy breaches and identity attack. Therefore, robust anonymization techniques are essential to safeguard patient confidentiality while preserving the clinical utility of the images.

With the rise of generative models like Generative Adversarial Networks (GANs), new approaches have emerged that leverage the generative power of these networks to tackle anonymization by creating synthetic data. This approach ensures that while the synthetic data retains key statistical properties and utility for analysis, it no longer corresponds to any real individual, offering a robust solution for protecting sensitive information. For instance, Jeon et al. \cite{jeon2022k} introduced $k$-SALSA, a framework utilizing a GAN-based approach to synthesize a $K$-anonymous dataset from private retinal images. However, the process of sample aggregation in the latent space comes with the drawback of reducing the dataset size by a factor of $k$. To address this issue, Pennisi et al. \cite{pennisi2023privacy} introduced a latent space navigation strategy to generate a diverse range of images. However, due to the weak supervision in the navigation process, the clinical interpretability of the generated images remains limited.

Unlike existing generative medical image anonymization methods that typically rely on GAN inversion approach ("$\E$-training"), where the encoder is trained with a fixed pre-trained StyleGAN generator to discover the latent representation of an input image, we approach the learning of the latent code as an image reconstruction task. This involves using a custom encoder and a co-training scheme to achieve a precise projection of the input image in its latent space.  Furthermore, we adopt the latent code optimization technique from \cite{barattin2023attribute} to explicitly manage the trade-off between identity protection and data utility, incorporating deep identity and utility losses tailored specifically for medical images.

\section{Proposed Method}
\begin{figure*}[htbp]
\centering
\includegraphics[width=0.8\textwidth]{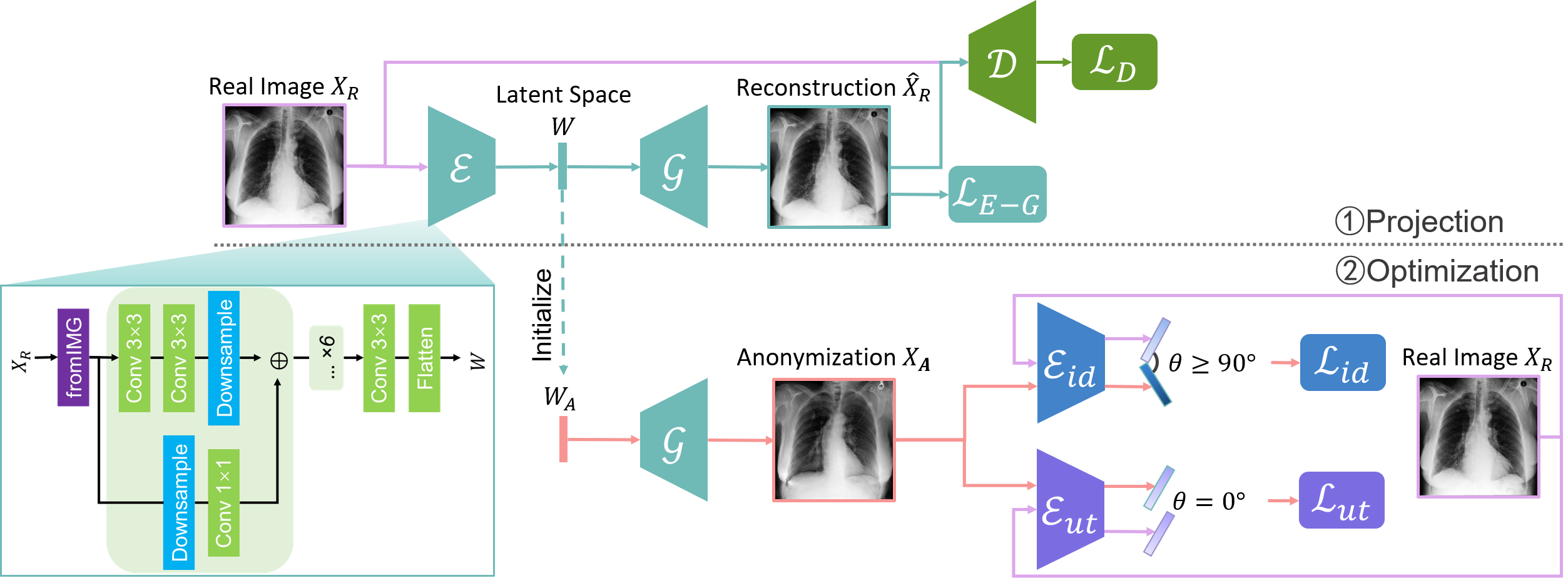}
\caption{Overview of the proposed method, consisting of two key stages: (1) AE-GAN network for latent code projection, and (2) Latent code optimization using identity removal loss and utility-preserving losss.
}
\label{fig1}
\end{figure*}

Given a private real image dataset, we aim to generate an anonymized version that ensures patient privacy while preserving data utility. 
As shown in Fig.~\ref{fig1}, we first project the real image into its latent space $W$ through image reconstruction. Next, we initialize the anonymized latent code $W_A$ as $W$ and optimize it using two loss functions: an identity loss $\Loss_{id}$ to obscure identifiable features, and a utility loss $\Loss_{ut}$ to retain critical diagnostic information.

\subsection{Latent Code Projection}
We formulate the latent code projection as an image reconstruction task using an AE-GAN (Autoencoder-Generative Adversarial Network) architecture (see top in Fig.~\ref{fig1}), which combines a custom encoder $\E$, a StyleGAN2 \cite{karras2020analyzing} generator $\G$ and discriminator $\D$. The encoder transforms an real image $X_R$ into its latent code $W=\E(X_R)$, while the generator reconstructs an approximation of the real image, $\hat{X}_R=\G(W)$ , from the latent code. A discriminator $\D(X_R)$ is also employed to distinguish between real and generated images.

The custom encoder $\E$ (lower left of Fig.~\ref{fig1}) mirrors the generator $\G$, leveraging the inherent symmetry of the AE architecture. $\E$ begins with an operation (fromIMG) that converts the input image into feature maps for subsequent convolutional processing. At each resolution, we implement a residual network architecture, consisting of two 3×3 convolution layers followed by a bilinear downsampling operation to reduce the feature map size. The skip connection includes a bilinear downsampling operation and a 1×1 convolution layer. After 6 operation groups, $\E$ projects the input image into the $W$ latent space of StyleGAN2 \cite{karras2020analyzing}. 

To learn a more accurate mapping from the input image to its \( W \) latent space, we propose a co-training scheme to optimize the AE-GAN network. In each training step, the encoder \(\E\) and generator \(\G\) are jointly optimized using a composite loss function \(\mathcal{L}_{\E, \G}\), which integrates three key components: pixel-wise similarity loss, perceptual similarity loss, and adversarial (generator) loss. This is followed by optimizing the discriminator \(\D\) using the discriminator loss \(\mathcal{L}_\D\).



\subsection{Latent Code Optimization}
Latent code optimization (see bottom right in Fig.~\ref{fig1}) drives the anonymization process, with $W_A$ as the only trainable parameter and initialized from $W$, while all other networks remain pre-trained and fixed. $W_A$ is optimized using two deep loss functions: $\Loss_{id}$ , which ensures that identifiable features are effectively obfuscated or removed, and $\Loss_{ut}$, which preserves critical visual and diagnostic information.

The identity loss $\Loss_{id}$ ensures that $X_A$ has a different identity from $X_R$, up to a desired margin. It is defined as follows:
\begin{equation}
\Loss_{id}(X_R,X_A)=\mathrm{max}(0, \mathrm{cos}(\E_{id}(X_R),\E_{id}(X_A))-m)
\end{equation}
where $\mathrm{cos}(a,b)=\frac{a \cdot b}{\mathrm{max}(||a||_2 \cdot ||b||_2,\epsilon)}$ denotes the cosine similarity, and $\epsilon$ is a small value to avoid division by zero. $\E_{id}$ is the identity encoder, which is pre-trained for identity recognition \cite{deng2019arcface}. $m$ is a hyperparameter that controls the dissimilarity between the real and the anonymized images, constrained to be equal to or greater than $\mathrm{arccos}(m)$. 

The utility loss $\Loss_{ut}$ enforces that utility attributes of $X_R$ are preserved in $X_A$. It is defined as follows:
\begin{equation}
\Loss_{ut}(X_R,X_A)=||\E_{ut}(X_R )-\E_{ut}(X_A)||_2
\end{equation}
where $\E_{ut}$ is the utility encoder, which is pre-trained to perform a downstream task (e.g. lung pathology classification). 

\section{Experiments}
\textbf{Dataset.} We perform anonymization on MIMIC-CXR-JPG dataset \cite{johnson2019mimic}, using the default dataset partition. To ensure sufficient data per patient, we apply a filtering strategy ($\geq$ 20 images per patient). This results in 50 875/522/1 641 images, corresponding to 1 538/14/47 patients in the training/validation/testing sets, respectively. The utility task focuses on classifying lung pathology labels, while the identity classifier is designed to predict patient IDs.

\noindent\textbf{Latent Code Projection.}
We evaluate the projection process using both the proposed "Co-training" scheme and the classic "$\E$-training" approach, comparing their performance in terms of image reconstruction fidelity.

\noindent\textbf{Utility Preservation.}
To assess the preservation of utility attributes, we train utility classifiers on the training sets of real images $X_R$, reconstructed images $\hat{X}_R$, and anonymized images $X_A$, then evaluate them on the same unseen real test set.

\noindent\textbf{Identity Anonymization.}
To measure the privacy-preserving properties of our approach, we first assess the singling-out risk\footnote{Singling out: the possibility to isolate some or all records which identify an individual in the dataset.} on testing set with two metrics: LC (local cloaking) and HR (hidden rate). LC is defined as the median number of images between a real image and its anonymization. HR is defined as the percentage of individuals in the real dataset whose anonymized images is not the most similar to them. 

Then, we introduce a protocol to assess linkability risk\footnote{Linkability: the ability to link, at least, two records concerning the same data subject or a group of data subjects.} from two perspectives: \textbf{Inner risk:} Verifying whether two anonymized images belong to the same patient. \textbf{Outer risk:} Verifying whether a real and its anonymized image belong to the same patient.
Specifically, we frame the evaluation as a standard identity verification task, where the identity classifier is trained on the real dataset to determine whether two images belong to the same patient. For the inner risk, the test set consists of image pairs from the anonymized test set. For the outer risk, the test set comprises image pairs, each containing a real image and its anonymized version from the test set.

\begin{table}[!htbp]
\caption{The results of image recontruction. 'PSNR': peak signal-to-noise ratio, 'SSIM': structural similarity, 'IW\_SSIM': information content weighted structural similarity, 'LPIPS': learned perceptual image patch similarity.}\label{tab1}
\centering
\begin{tabular}{l|cccc}
\hline
Method &PSNR$\uparrow$ &SSIM$\uparrow$ &IW\_SSIM$\uparrow$ &LPIPS$\downarrow$\\
\hline
Co-training &24.319 &0.968 &0.602 &0.340\\
$\E$-training &19.477&0.936 &0.363 &0.446\\
\hline
\end{tabular}
\end{table}
Lastly, we perform membership inference attacks (MIA) \cite{shokri2017membership} to predict whether an anonymized image was part of the real training set. The attacker is trained on the real dataset and evaluated on the anonymized dataset (target dataset). 

\section{Results}
\begin{figure}[!htb]
	\centering
	\subfigure{
	\rotatebox{90}{\scriptsize{~~~~~~~$X_R$}}
		\begin{minipage}[t]{0.95\linewidth}
			\centering
			\includegraphics[width=1\linewidth]{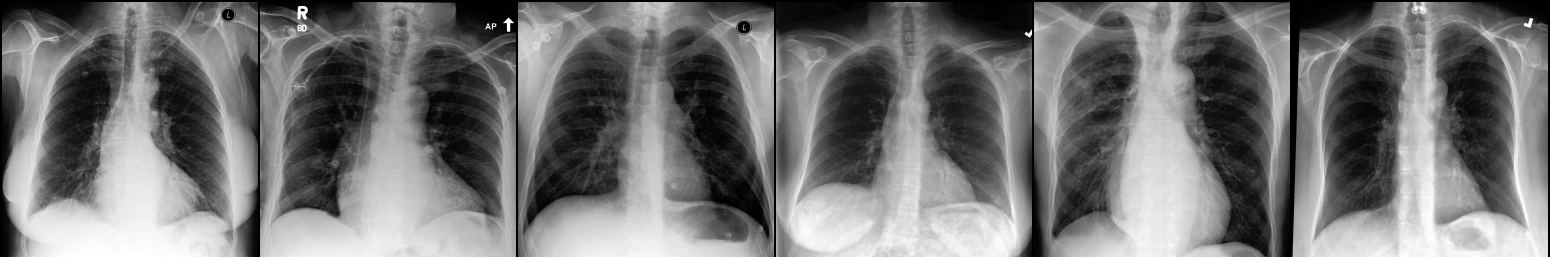}
		\end{minipage}
	}\vspace{-0.5\baselineskip}
	\subfigure{
	\rotatebox{90}{\scriptsize{~Co-training}}
		\begin{minipage}[t]{0.95\linewidth}
			\centering
			\includegraphics[width=1\linewidth]{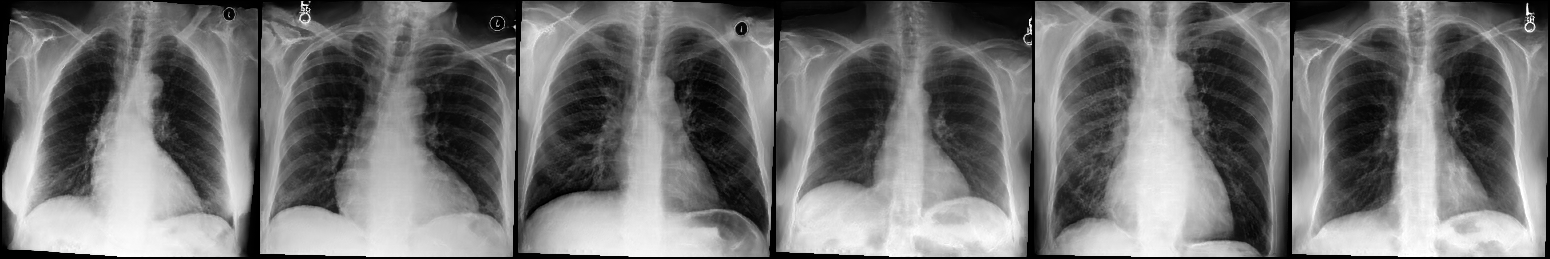}
		\end{minipage}
	}\vspace{-0.5\baselineskip}
	\subfigure{
	\rotatebox{90}{\scriptsize{~$\E$-training}}
		\begin{minipage}[t]{0.95\linewidth}
			\centering
			\includegraphics[width=1\linewidth]{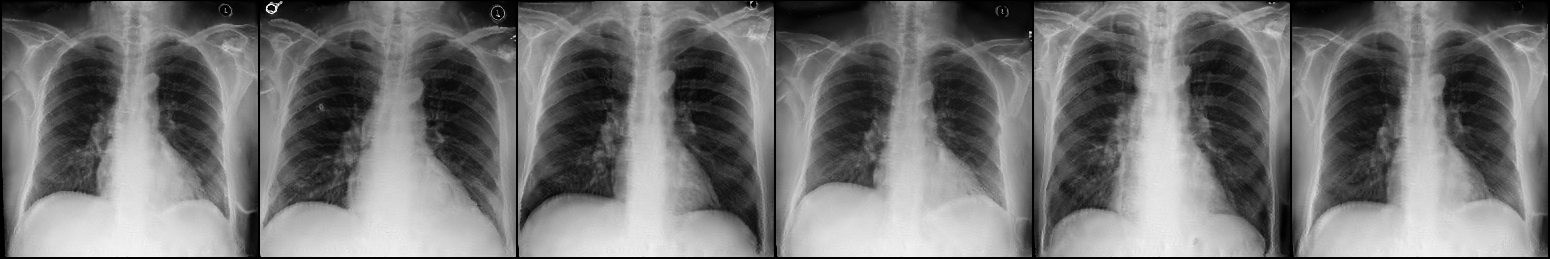}
		\end{minipage}
	}
	\caption{Reconstruction results. The first row displays the real images $X_R$. The last two rows show the reconstructed images $\hat{X}_R$ produced by the proposed co-training scheme and the $\E$-training scheme, respectively.
	}
	\label{fig2}
\end{figure}

\textbf{Image Reconstruction.} Table \ref{tab1} shows that the co-training scheme outperforms the \(\E\)-training approach in reconstruction accuracy, as indicated by higher PSNR and IW\_SSIM scores, highlighting the benefits of joint optimization for enhancing reconstruction quality. The results are further validated by the visual comparison in Fig.~\ref{fig2}, where the co-training scheme recovers input images with finer details and higher fidelity. In contrast, the $\E$-training scheme exhibits noticeable discrepancies when compared to the original images.

\noindent\textbf{Utility Preservation.} As shown in Table \ref{tab2}, the performance of classifiers trained on anonymized datasets (row 4-6) exhibits minimal difference compared to the one trained on real dataset (row 1), indicating that the anonymized datasets maintain a similar level of data utility compared to their real counterparts.
Moreover, incorporating $\Loss_{ut}$ (row 4, 6) yields better results compared to those without it (row 2, 5), highlighting the effectiveness of the utility loss $\Loss_{ut}$ in preserving utility-related information during optimization.
Additionally, classifier trained on reconstructed datasets using $\E$-training scheme (row 3) shows the weakest performance, highlighting the advantages of the proposed co-training scheme.

\begin{table}[!htbp]
\caption{Results of utility preservation. 'Acc': accuracy, 'AP': average precision, 'AUROC': area under the ROC curve, 'F1': F1 score.}\label{tab2}
\centering
\begin{tabular}{ll|cccc}
\hline
\multicolumn{2}{l|}{Training Data} &~Acc$\uparrow$~ &~AP$\uparrow$~ &AUROC$\uparrow$ &~~F1$\uparrow$~~\\
\hline
\multicolumn{2}{l|}{$X_R$} &0.740 &0.631 &0.774 &0.511\\

\hline 
\multirow{2}*{$\hat{X}_R$} &Co-training &0.718 &0.586 &0.737 &0.451\\
&$\E$-training &0.654 &0.472 &0.640 &0.307\\
\hline 
\multirow{3}*{$X_A$}
&$\Loss_{ut}$ &0.726 &0.600 &0.747 &0.493\\
&$\Loss_{id}$  &0.696 &0.559 &0.713 &0.424\\
&$\Loss_{ut}+\Loss_{id}$ &0.713 &0.578 &0.737 &0.531\\
\hline 
\end{tabular}
\end{table}
\noindent\textbf{Identity Anonymization.}
Table \ref{tab4} denotes that the anonymized images (row 3-5, 7-9) exhibit a lower linkability risk compared to their real counterparts (row 1). Additionally, the anonymized datasets exhibit high inner risk and a low outer risk, which is expected. The high inner risk indicates that the anonymized images retain relative likability relationships (e.g., images from the same patient remain associated with each other), while the low outer risk signifies that the linkability relationship between anonymized images and their real counterparts is effectively removed (e.g., anonymized images do not resemble their real counterparts in terms of identity).
\begin{table}[!htbp]
\caption{Results of linkability risk. 'Pre': pre-training, 'F1': F1 score, 'Acc': accuracy, 'TAR': true accept rate, 'FAR': false accept rate.}\label{tab4}
\centering
\begin{tabular}{l|ll|cccc}
\hline
&\multicolumn{2}{l|}{Testing Data} &F1$\uparrow$ &Acc$\uparrow$ &TAR$\uparrow$ &FAR$\downarrow$\\ 
\hline
&\multicolumn{2}{l|}{$X_R$} &0.983 &0.983 &0.966 &0.034\\
\hline
\multirow{4}*{\rotatebox{90}{Inner risk}} &\multirow{1}*{$\hat{X}_R$} &Co-training &0.611 &0.720 &0.441 &0.559\\
\cline{2-7}
&\multirow{3}*{$X_A$}
&$\Loss_{ut}$ &0.665 &0.748 &0.500 &0.500\\
&&$\Loss_{id}$ &0.855 &0.873 &0.746 &0.254\\
&&$\Loss_{ut}+\Loss_{id}$ &0.887 &0.899 &0.798 &0.202\\

\hline
\multirow{4}*{\rotatebox{90}{Outer risk}} 
&\multirow{1}*{$\hat{X}_R$} &Co-training &0.139 &0.075 &0.075 &0.925\\
\cline{2-7}
&\multirow{3}*{$X_A$}
&$\Loss_{ut}$ &0.136 &0.073 &0.073 &0.927\\
&&$\Loss_{id}$ &0.000 &0.000 &0.000 &1.000\\
&&$\Loss_{ut}+\Loss_{id}$ &0.000 &0.000 &0.000 &1.000\\
\hline
\end{tabular}
\end{table}

Moreover, incorporating $\Loss_{id}$ increases the inner risk while reducing the outer risk. This suggests the loss $\Loss_{id}$ helps push the anonymized images belonging to the same patient closer together while breaking the linkability to their real counterparts.
Similar results are shown in Table \ref{tab3}, where $\Loss_{id}$ plays a crucial role in reducing the singling out risk, as incorporating $\Loss_{id}$ (row 3-4) produces anonymized datasets with a low risk of singling out compared to the datasets without it (row 1-2).
\begin{table}[!htbp]
\caption{Results for singling out risk. '$N_R$': median number of real images, '$N_A$' median number of anonymized images.}\label{tab3}
\centering
\begin{tabular}{ll|ccc}
\hline
\multicolumn{2}{l|}{\multirow{2}*{Testing Data}} &\multirow{2}*{HR$\uparrow$} &\multicolumn{2}{c}{LC$\uparrow$}\\
\cline{4-5}
&&&~$N_R$~ &~$N_A$~\\ 
\hline
\multirow{1}*{$\hat{X}_R$} &Co-training &0.936 &41 &58\\
\hline
\multirow{3}*{$X_A$} &$\Loss_{ut}$ &0.936 &41 &56\\
&$\Loss_{id}$ &1.000 &1546 &1639\\
&$\Loss_{ut}+\Loss_{id}$ &1.000 &1529 &1639\\
\hline
\end{tabular}
\end{table}

In the MIA attack, attackers achieve high accuracy (Acc: 0.968) during training, demonstrating their effectiveness on the test dataset. However, during the evaluation stage, their performance significantly drops, with an accuracy of 0. This indicates a considerable challenge in correctly predicting the membership status of anonymized images.

\begin{figure}[htbp]
	\centering
    \vspace{-0.5\baselineskip}
	\subfigure{
	    \rotatebox{90}{\scriptsize{~~~~~~~~Train}}
		\begin{minipage}[t]{0.18\linewidth}
			\centering
			\includegraphics[width=1\linewidth]{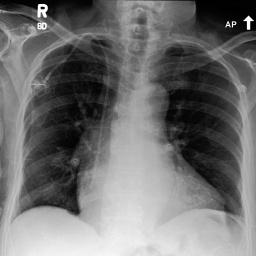}
		\end{minipage}
		\begin{minipage}[t]{0.18\linewidth}
			\centering
			\includegraphics[width=1\linewidth]{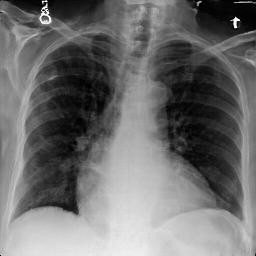}
		\end{minipage}
		\begin{minipage}[t]{0.18\linewidth}
			\centering
			\includegraphics[width=1\linewidth]{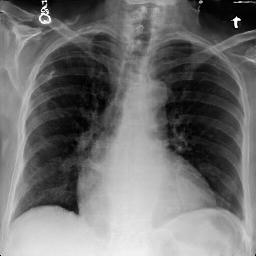}
		\end{minipage}
		\begin{minipage}[t]{0.18\linewidth}
			\centering
			\includegraphics[width=1\linewidth]{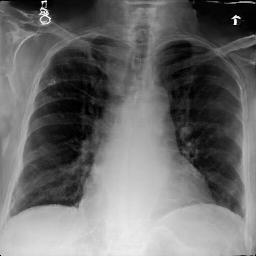}
		\end{minipage}
		\begin{minipage}[t]{0.18\linewidth}
			\centering
			\includegraphics[width=1\linewidth]{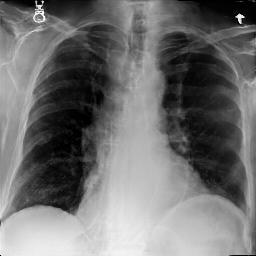}
		\end{minipage}
	}\vspace{-0.5\baselineskip}
	\subfigure{
	    \rotatebox{90}{\scriptsize{~~~~~~~~~Valid}}
		\begin{minipage}[t]{0.18\linewidth}
			\centering
			\includegraphics[width=1\linewidth]{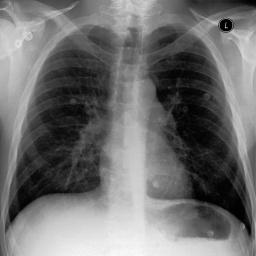}
		\end{minipage}
		\begin{minipage}[t]{0.18\linewidth}
			\centering
			\includegraphics[width=1\linewidth]{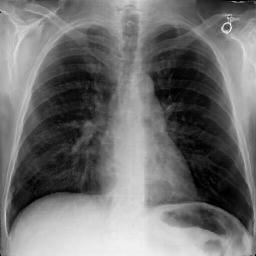}
		\end{minipage}
		\begin{minipage}[t]{0.18\linewidth}
			\centering
			\includegraphics[width=1\linewidth]{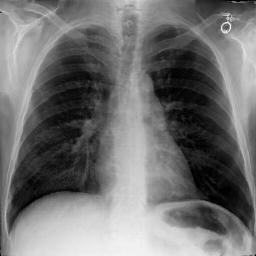}
		\end{minipage}
		\begin{minipage}[t]{0.18\linewidth}
			\centering
			\includegraphics[width=1\linewidth]{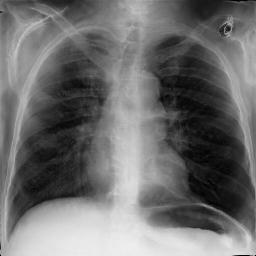}
		\end{minipage}
		\begin{minipage}[t]{0.18\linewidth}
			\centering
			\includegraphics[width=1\linewidth]{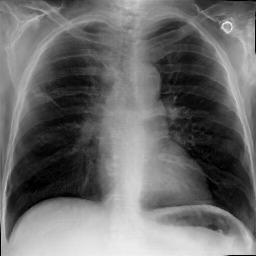}
		\end{minipage}
	}\vspace{-0.5\baselineskip}
    \setcounter{subfigure}{0}
	\subfigure{
	    \rotatebox{90}{\scriptsize{~~~~~~~~~Test}}
		\begin{minipage}[t]{0.18\linewidth}
			\centering
			\includegraphics[width=1\linewidth]{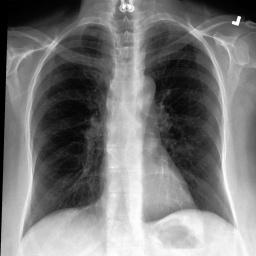}
			\footnotesize{$(a) X_R$}
		\end{minipage}
		}\hspace{-0.45\baselineskip}
	\subfigure{
		\begin{minipage}[t]{0.18\linewidth}
			\centering
			\includegraphics[width=1\linewidth]{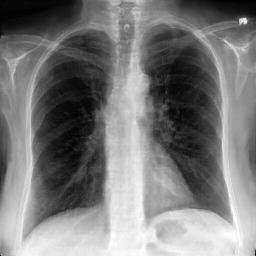}
			\footnotesize{$(b) \hat{X}_R$}
		\end{minipage}
		}\hspace{-0.45\baselineskip}
	\subfigcapskip=0.4\baselineskip
	\subfigure{
		\begin{minipage}[t]{0.18\linewidth}
			\centering
			\includegraphics[width=1\linewidth]{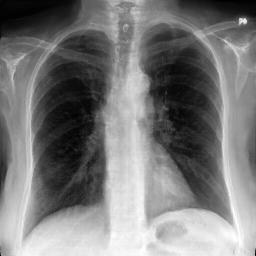}
			\footnotesize{$(c) X_A: \Loss_{ut}$}
		\end{minipage}
		\begin{minipage}[t]{0.18\linewidth}
			\centering
			\includegraphics[width=1\linewidth]{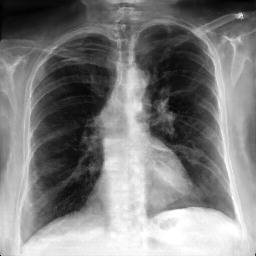}
			\footnotesize{$\Loss_{id}$}
		\end{minipage}
		\begin{minipage}[t]{0.18\linewidth}
			\centering
			\includegraphics[width=1\linewidth]{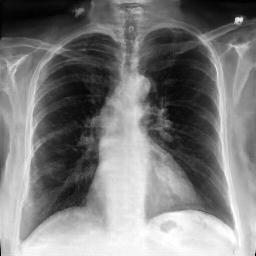}
			\footnotesize{$\Loss_{ut}+\Loss_{id}$}
		\end{minipage}
	}\vspace{-0.5\baselineskip}
	\caption{Anonymization results. Real images $X_R$ randomly selected from the training, validation, and test sets are displayed in the first column. The corresponding reconstructed images $\hat{X}_R$ are displayed in the second column. The anonymized images $X_A$ are displayed in the last three columns.
	}
	\label{fig3}
\end{figure}
\noindent\textbf{Qualitative Results}
The visualization results are shown in Fig.~\ref{fig3}. In this figure, the anonymized images optimized using only the utility loss \(\Loss_{ut}\) (column 3) exhibit greater visual similarity to their real counterparts (column 1), while those optimized using only the identity loss \(\Loss_{id}\) (column 4) appear more distinct from their originals. Additionally, the anonymized images optimized with both the $\Loss_{ut}$ and the $\Loss_{id}$ (column 5) strike a balance, appearing more realistic by simultaneously considering both identity removal and utility preservation.

\section{Conclusions}
In this paper, we introduce an innovative two-stage approach to address medical image anonymization. First, we precisely project the input image into its latent space using image reconstruction with a custom encoder and an effective co-training scheme. Then, we optimize the latent code to remove identifiable information while preserving utility using two deep loss functions. Our results demonstrate that the method effectively anonymizes the images while better preserving utility attributes, leading to improved de-identification and retention of critical clinical information.

\blfootnote{\textbf{Acknowledgements.} This work has been supported by the French government,
through the 3IA Côte d’Azur Investments in the Future project managed by the National Research Agency (ANR) with the reference number ANR-19-P3IA-0002, and supported by the Inria Sophia Antipolis - Méditerranée, "NEF" computation cluster. }

\bibliographystyle{IEEEbib}
\bibliography{refs}

\end{document}